\documentclass[review]{elsarticle}

\usepackage{amssymb}
\usepackage{amsmath, amssymb}
\usepackage{makecell}
\usepackage{multirow}
\usepackage{booktabs}

\usepackage{subfig}
\usepackage{graphicx}

\newcommand{\etal}{\textit{et al}.}
\newcommand{\ie}{\textit{i}.\textit{e}.}
\newcommand{\eg}{\textit{e}.\textit{g}.}

\journal{Pattern Recognition}

\begin{document}

\begin{frontmatter}



\title{Video Semantic Segmentation with Inter-Frame Feature Fusion and Inner-Frame Feature Refinement}




\author{Jiafan Zhuang}
\ead{jfzhuang@mail.ustc.edu.cn}
\author{Zilei Wang\corref{fn1}}
\ead{zlwang@ustc.edu.cn}
\author{Junjie Li}
\ead{hnljj@mail.ustc.edu.cn}

\cortext[fn1]{Corresponding author.}

\address{National Engineering Laboratory for Brain-inspired Intelligence Technology and Application, University of Science and Technology of China, Hefei 230027, China}

\begin{abstract}
Video semantic segmentation aims to generate accurate semantic maps for each video frame. To this end, many works dedicate to integrate diverse information from consecutive frames to enhance the features for prediction, where a feature alignment procedure via estimated optical flow is usually required. However, the optical flow would inevitably suffer from inaccuracy, and then introduce noises in feature fusion and further result in unsatisfactory segmentation results. 
In this paper, to tackle the misalignment issue, we propose a spatial-temporal fusion (STF) module to model dense pairwise relationships among multi-frame features. Different from previous methods, STF uniformly and adaptively fuses features at different spatial and temporal positions, and avoids error-prone optical flow estimation.
Besides, we further exploit feature refinement within a single frame and propose a novel memory-augmented refinement (MAR) module to tackle difficult predictions among semantic boundaries. Specifically, MAR can store the boundary features and prototypes extracted from the training samples, which together form the task-specific memory, and then use them to refine the features during inference. Essentially, MAR can move the hard features closer to the most likely category and thus make them more discriminative.
We conduct extensive experiments on Cityscapes and CamVid, and the results show that our proposed methods significantly outperform previous methods and achieves the state-of-the-art performance.
Code and pretrained models are available at https://github.com/jfzhuang/ST\_Memory.
\end{abstract}




\end{frontmatter}


\section{Introduction}
\label{sec:introduction}
Semantic segmentation targets to assign each pixel in scene images a semantic class, which is one of the fundamental tasks in computer vision. In recent years, image semantic segmentation has achieved unprecedented performance benefited from the great progress of deep convolutional neural network (DCNN)~\cite{long2015fully} and construction of various datasets (\eg Cityscapes~\cite{cordts2016cityscapes} and CamVid~\cite{brostow2009semantic}). However, many real-world applications have strong demands for accurate video semantic segmentation, \eg, robotics, autonomous driving, and video surveillance. Actually, video data offer richer information than static images, \eg, diverse presentations from multiple frames and temporal consistency prior. Thus video can provide good potential to achieve more accurate semantic segmentation. The key is how to produce more discriminative features by exploiting the characteristics of videos.

A natural way to enhance video features is to integrate the diverse information of consecutive frames~\cite{gadde2017semantic,nilsson2018semantic}. Specifically, the feature alignment is commonly performed via the optical flow based feature warping, which ensures that pixel-level features at the same spatial position represent the identical object, and then the temporal feature fusion is conducted for each pixel. Evidently, the accurate optical flow is critical for feature fusion.  However, the optical flow estimation inevitably suffers from inaccuracy in the boundary areas due to object occlusion and plain texture~\cite{liu2019selflow,zhuang2020video}. If the features are not well-aligned, the noises would be introduced. Consequently, the quality of fused features would be reduced and the segmentation performance would be deteriorated.

Besides, after aggregating information from consecutive frames, can we further refine the fused feature? Different from inter-frame feature fusion in video segmentation methods, some image-based methods adopt the post-processing techniques to optimize the features for prediction. For example, DenseCRF~\cite{krahenbuhl2011efficient} uses a graph structure to model pairwise potentials on all pixels and iteratively adjusts the feature by optimizing an energy function. Essentially, it uses similar features to mutually enhance themselves. SegFix~\cite{yuan2020segfix} proposes to replace the difficult boundary features with some better ones, whose locations are predicted by a network and often lie around the boundary areas in practice.

\begin{figure*}[t]
	\begin{center}
		\includegraphics[width=1.0\linewidth]{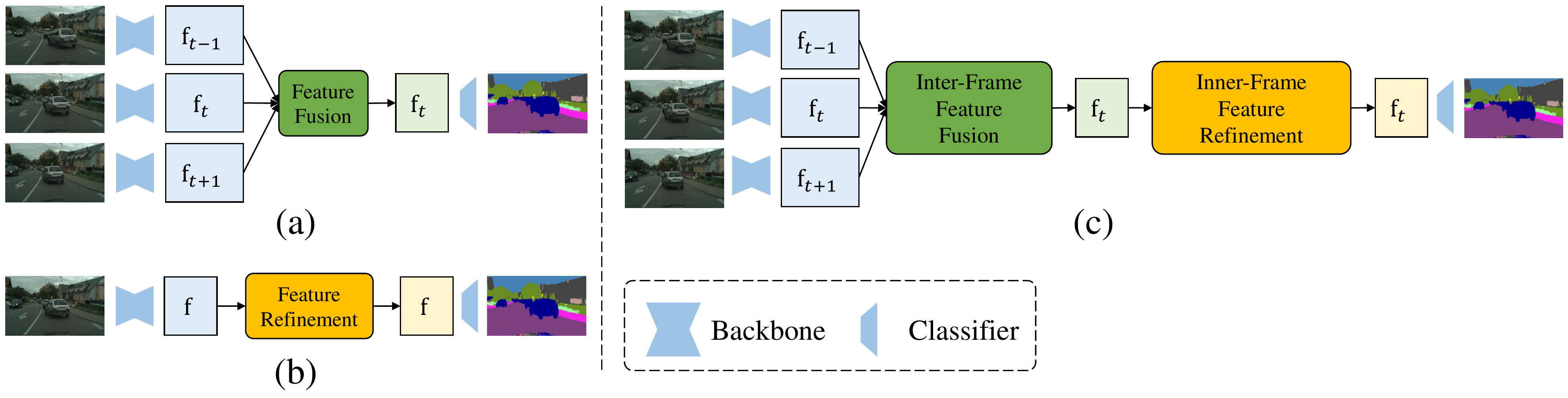}
	\end{center}
	\caption{\textbf{Architecture illustrations for different methods}. (a) Feature fusion in video segmentation methods (\eg, NetWarp~\cite{gadde2017semantic} and GRFP~\cite{nilsson2018semantic}). (b) Feature refinement in image segmentation methods (\eg, DenseCRF~\cite{krahenbuhl2011efficient} and SegFix~\cite{yuan2020segfix}). (c) Our proposed method. Best viewed in color.}
	\label{intro}
\end{figure*}

Actually, feature fusion is proposed to aggregate useful information from different frames while feature refinement is designed for correcting error-prone features, which are potentially complementary.
Based on this motivation, in this paper, we aim to improve the accuracy of video semantic segmentation by simultaneously considering inter-frame feature fusion and inner-frame feature refinement, as shown in Figure~\ref{intro}.
For the inter-frame fusion, we need to tackle the feature misalignment issue. To this end, we propose a spatial-temporal fusion (STF) module that uniformly fuses the features at different spatial and temporal positions and does not require explicit feature alignment via error-prone optical flows. Here the transformer~\cite{vaswani2017attention} is particularly adopted due to the power to model long-range dependencies. To be specific, the encoder is fed with the features extracted from consecutive frames, and the decoder is used to generate the prediction features by retrieving the current frame from the encoded features. In particular, we utilize the self-attention mechanism in transformer to guide the feature fusion in latent space, in which more similar features are supposed to be more likely to represent the same object. For an image pixel, hence, STF would integrate multiple similar features at different temporal and spatial positions, rather than only the temporal-aligned features in the previous works~\cite{gadde2017semantic,nilsson2018semantic}.

In addition, an image with the resolution of $(1024, 2048)$ would typically produce the features with the resolution of $(128, 256)$. The transformer taking three frames needs to process $3 \times 128 \times 256=98304$ pixel-level features, which results in unacceptable computation and memory cost with $\mathcal{O}(N^{2})$ complexity when computing affinity matrix. Inspired by a recent work~\cite{huang2019interlaced}, we propose an interlaced cross-self attention (ICSA) attention mechanism to divide the dense affinity matrix computation in transformer as the product of a long-range cross-attention and a short-range self-attention, which can greatly reduce the memory consumption. 


On the other hand, we propose inner-frame feature refinement to further adjust the fused features for better prediction without devising more complicated network structure. In this work, we focus on refining the hard features that are error-prone and always lie in the boundary areas of different classes~\cite{yuan2020segfix}. To this end, we propose a novel memory-augmented refinement (MAR) module that uses the stored features in memory to augment the hard features. Actually, this is motivated by an intuitive observation that humans would retrieve memory to enhance the judgement when facing semantically ambiguous contents.
Here the memory represents the experience from the training samples. For each semantic category,  we particularly store the hard features and their corresponding class prototypes (refer to the mean feature representing a single category), which together form a key-value memory bank. 
During inference, a hard feature would be refined by the class prototypes, where the weights of different classes are computed by comparing it with the stored hard features in the memory. 
In this way, the discriminativeness of boundary features would be enhanced since MAR would make them move closer to the most likely category. Evidently, MAR has good interpretability and can be conveniently inserted into different models as an independent module.

We experimentally evaluate the proposed method on the Cityscapes and CamVid datasets. The results validate the effectiveness of our STF and MAR to improve the quality of features, and their combination can achieve the state-of-the-art segmentation performance.

The contributions of this work are summarized as
\begin{itemize}
	\item We design a novel video semantic segmentation framework by simultaneously considering inter-frame feature fusion and inner-frame feature refinement, which can take advantages from both two feature enhancement techniques and effectively improve segmentation accuracy.
	\item We propose an effective spatial-temporal fusion module based on the transformer, which can uniformly aggregate the features at different spatial and temporal positions and avoid error-prone temporal feature alignment.
	\item We propose a novel memory-augmented refinement module to particularly refine hard features using the experience from training samples. In particular, the key-value memory is stored to refine the hard features closer to the most likely category.
	\item We experimentally evaluate the effectiveness of our proposed methods, and the results on Cityscapes and CamVid demonstrate the superiority of our methods to previous state-of-the-art methods.
\end{itemize}

The rest of this paper is organized as follows. We review the related works on image and video semantic segmentation, transformer and memory mechanism in Section~\ref{sec:related}. Section~\ref{sec:approach} provides the details of our approach, and Section~\ref{sec:exp} experimentally evaluates the proposed method. Finally, we conclude the work in Section~\ref{sec:conclusion}.

\section{Related Work}
\label{sec:related}

\subsection{Image Semantic Segmentation}
With the development of DCNN, more semantic segmentation networks spring up. Specifically, the fully convolutional networks (FCNs)~\cite{long2015fully} firstly uses the convolutional layers to replace fully-connected layers and can achieve better performance. Inspired by FCN, many extensions have been proposed to advance image semantic segmentation. The dilated layers~\cite{chen2017deeplab} are used to replace the pooling layers, which can better balance the computational cost and size of receptive fields. To further improve segmentation accuracy, spatial pyramid pooling and atrous spatial pyramid pooling (ASPP) are used in PSPNet~\cite{zhao2017pyramid} and DeepLab~\cite{chen2017deeplab} to capture multi-scale contextual information.
Mitivated by ASPP, Peng~\etal~\cite{PENG2020107498} proposes a stride spatial pyramid
pooling (SSPP) to capture multiscale semantic information from the high-level feature map, while Lian~\etal~\cite{LIAN2021107622} proposes a cascaded hierarchical atrous pyramid pooling module to simultaneously extract rich local detail characteristics and important global contextual information.
CENet~\cite{ZHOU2022108290} aggregates contextual cues via densely usampling the convolutional features of deep layer to the shallow deconvolutional layers, which can fully explore multiple scale contextual information.
GPNet~\cite{ZHANG2021107940} densely captures and filters the multi-scale information in a gated
and pair-wise manner with a gated pyramid module and a cross-layer attention module. 
Marin~\etal~\cite{ORSIC2021107611} propose a novel architecture based on shared pyramidal representation and fusion of heterogeneous features along the upsampling path, which is effective for dense inference in images with large scale.
Different from enhancing features, EFNet~\cite{WANG2021108023} propose to produce multiple enhanced images and fuses them to yield one new image, which can encourage the model to exploit complementary information.

Differently, our proposed methods focus on exploiting both spatial and temporal contexts to further improve the performance and can build upon any existing image segmentation models.

\subsection{Video Semantic Segmentation}

Different from static images, videos embody rich temporal information that can be exploited to improve the semantic segmentation performance. Existing video semantic segmentation methods mainly fall into two categories. The first category aims to accelerate inference speed by reusing the features in previous frames. DFF~\cite{zhu2017deep} estimates the optical flow fields~\cite{ilg2017flownet} from the key frame to other frames and then propagates the high-level features using the predicted optical flows. Accel~\cite{jain2019accel} proposes a reference branch to extract high-quality segmentation from the key frames and an update branch to efficiently extract low-quality segmentation from the current frames, and the fuses them to improve the segmentation accuracy. DAVSS~\cite{zhuang2020video} designs a feature correction mechanism to tackle distorted features after propagation due to inaccurate optical flows. LERNet~\cite{WU2020107268} proposes to propagate multi-level features from the key frame via a temporal holistic attention module.
TDNet~\cite{hu2020temporally} distributes several sub-networks over sequential frames and then recomposes the extracted features for segmentation via an attention propagation module. Differently, Liu~\etal~\cite{liu2020efficient} designs a new temporal knowledge distillation methods to narrow the performance gap between compact models and large models.

Another category focus on improving segmentation accuracy by modeling cross-frame relations to integrate information from consecutive frames.  
V2V~\cite{tran2016deep} utilizes a 3D CNN to perform a voxel-level prediction. STFCN~\cite{fayyaz2016stfcn} utilizes a spatial-temporal LSTM over per-frame CNN features. However, these methods cannot achieve high performance due to rough processing of different frames.
HDCNN~\cite{WANG2017437} proposes a transition layer structure to make the pixel-wise label prediction consist with adjacent pixels across space and time domains.
Recently, some works~\cite{gadde2017semantic,nilsson2018semantic} propose to fuse features from multiple frames to produce the better features for prediction. They usually adopt the optical flow to model cross-frame relationships and perform temporal alignment by warping features. In particular, NetWarp~\cite{gadde2017semantic} uses a set of learnable weights to fuse multiple features, and GRFP~\cite{nilsson2018semantic} proposes the gated recurrent units STGRU to estimate the uncertainty of warped features and then conducts feature fusion on the areas with high reliability. Obviously, the optical flow is critical for feature alignment and would affect the final accuracy. However, the optical flow estimation inevitably suffers from inaccuracy, especially for the occlusion areas and small objects (\eg, pedestrian, pole)~\cite{zhuang2020video}. 
In this work, we follow the route of second category and focus on improving segmentation accuracy. Different from previous works, we propose to simultaneously model the spatial-temporal relationship without feature alignment, which can avoid error-prone optical-flow estimation. Furthermore, we propose to use memory to refine the prediction features.

\subsection{Transformer}

Transformer is originally proposed for the sequence-to-sequence machine translation~\cite{vaswani2017attention}, and currently has dominated various NLP tasks. As the core component of transformer, the self-attention is particularly suitable for modeling long-range dependencies. Due to the success of transformer in the NLP field, some works attempt to explore the benefits of transformer in computer vision. 
DETR~\cite{carion2020end} first builds an object detection system based on transformer, which can reason about relationships between objects and global context and directly output the final set of predictions.
Swin Transformer~\cite{liu2021swin} designs a novel shifted windowing scheme, which can limit attention computation to local windows while also allow for cross-window connection. It achieves an impressive performance on a broad range of vision tasks.
In this paper, we propose STF by using transformer to model the spatial-temporal relationship among pixel-wise features extracted from consecutive frames. To our best knowledge, this is the first attempt to exploit the transformer in video semantic segmentation.

Recently, Action Transformer~\cite{girdhar2019video} and Actor Transformer \cite{gavrilyuk2020actor} also adopt transformer to model spatial-temporal relationship in action detection and group action recognition tasks, which are closely related to our proposed STF. 
They naturally adopt transformer for modeling proposal-context and proposal-proposal relationship. But our proposed STF is different from these two works. STF is designed for modeling pixel-wise relationship, which would involve huge memory and computation overhead issues. In this work, we propose interlaced cross-self attention (ICSA) mechanism to tackle these issues and achieve efficient global relationship modeling.

\subsection{External Memory}

In DCNN, external memory is generally used to enhance feature representations by storing history data, which is especially useful for the tasks without enough samples, \eg, life-long learning~\cite{NEURIPS2019_f8d2e80c} and few-shot learning~\cite{ZHANG2020107348,cai2018memory}. 
For example, MM-Net~\cite{cai2018memory} proposes to store the representative features in the support set for one-shot learning, and then use them to predict the parameters of feature extraction network on query images. Actually, this can make the query features more relevant to the support features.
In recent years, the memory mechanism is also exploited to store long-range temporal contexts for video tasks during inference. 
In video object detection, \cite{deng2019object} proposes to store pixel-level and instance-level features extracted from previous frames and then use them to enhance the current frame.
LFB~\cite{wu2019long} proposes a long-term feature bank for action localization to store supportive information extracted over the entire span of a video, and then uses them to enhance the short-term features extracted from short video clips. 
Different from the previous works that store temporal~\cite{deng2019object,wu2019long} or sample~\cite{cai2018memory} contexts, in this paper we propose to store the hard features and class prototypes from the training samples to form a task-specific memory, and then use them to refine the boundary features during inference.

\section{Our Approach}
\label{sec:approach}
In this work, we aim to boost the accuracy of video semantic segmentation by enhancing the features for prediction. To this end, we first propose a spatial-temporal fusion (STF) module to perform inter-frame feature fusion at different spatial and temporal positions, which can avoid error-prone optical flow estimation. Then we propose a memory-augmented refinement (MAR) module to further refine the boundary features during inference, which is essential to utilize the stored experience from training samples. 
In the following, we first introduce the framework of our proposed  approach, and then elaborate on two key modules, namely, STF and MAR.

\subsection{Framework}

\begin{figure*}[t]
	\begin{center}
		\includegraphics[width=1.0\linewidth]{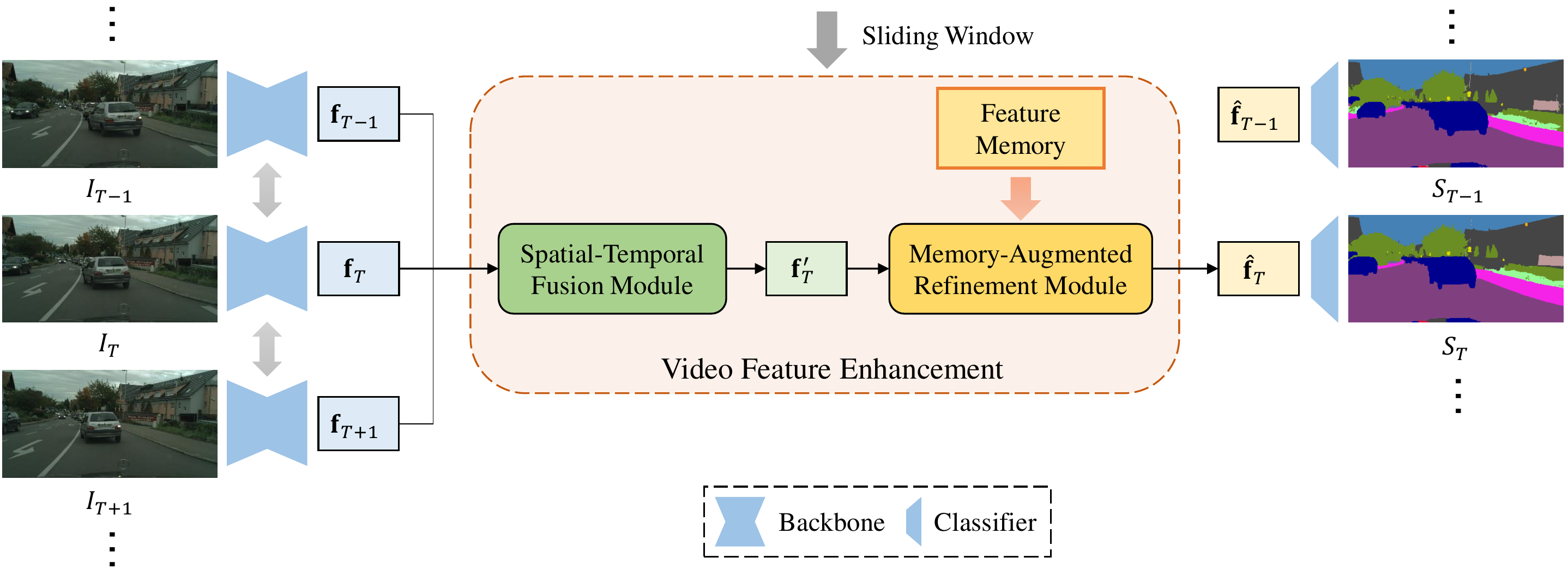}
	\end{center}
	\caption{\textbf{The framework of our proposed approach}. First, the feature is extracted by an image segmentation model for each frame. Then the features of consecutive frames are fed into our proposed STF module to perform feature fusion. After that, the fused feature $\mathbf{f}_{T}^{'}$ is further refined by our proposed MAR module, resulting in $\widehat{\mathbf{f}}_{T}$. Finally, the segmentation result is obtained by applying the classifier on $\widehat{\mathbf{f}}_{T}$. Best viewed in color.}
	\label{framework}
\end{figure*}

Our proposed video semantic segmentation framework is illustrated in Figure.~\ref{framework}. Formally, given a sequence of $n$ video frames denoted by \{$I_1$, $I_2$, $\cdots$, $I_n$\}, our purpose is to get the accurate semantic segmentation maps for every video frame, denoted by \{$S_1$, $S_2$, $\cdots$, $S_n$\}. Specifically, we first extract features from each frame image using an off-the-shelf segmentation model. Then we conduct video feature enhancement for the current timestamp $T$ with a sequence of three-frame features \{$\mathbf{f}_{T-1}$, $\mathbf{f}_{T}$, $\mathbf{f}_{T+1}$\}, resulting in $\widehat{\mathbf{f}}_{T}$ for final prediction. Finally, we apply the classifier on $\widehat{\mathbf{f}}_{T}$ to produce the segmentation result $S_{T}$. Since such a procedure can be performed in a sliding-window manner, we can obtain the corresponding segmentation sequence.

In this work, we dedicate to enhance video features to improve the segmentation performance. To be specific, we first feed the sequence of frame features into our proposed STF module to capture spatial-temporal dependencies and complete pixel-wise feature fusion, resulting in the fused feature $\mathbf{f}_{T}^{'}$. After that, our proposed MAR module further refines $\mathbf{f}_{T}^{'}$ into $\widehat{\mathbf{f}}_{T}$ by exploiting the stored feature memory to enhance the discriminativeness of the boundary features. Evidently, STF and MAR are the key components of our method that determine the performance of video semantic segmentation.

\subsection{Spatial-Temporal Fusion}

\begin{figure}[t]
	\begin{center}
		\includegraphics[width=0.5\linewidth]{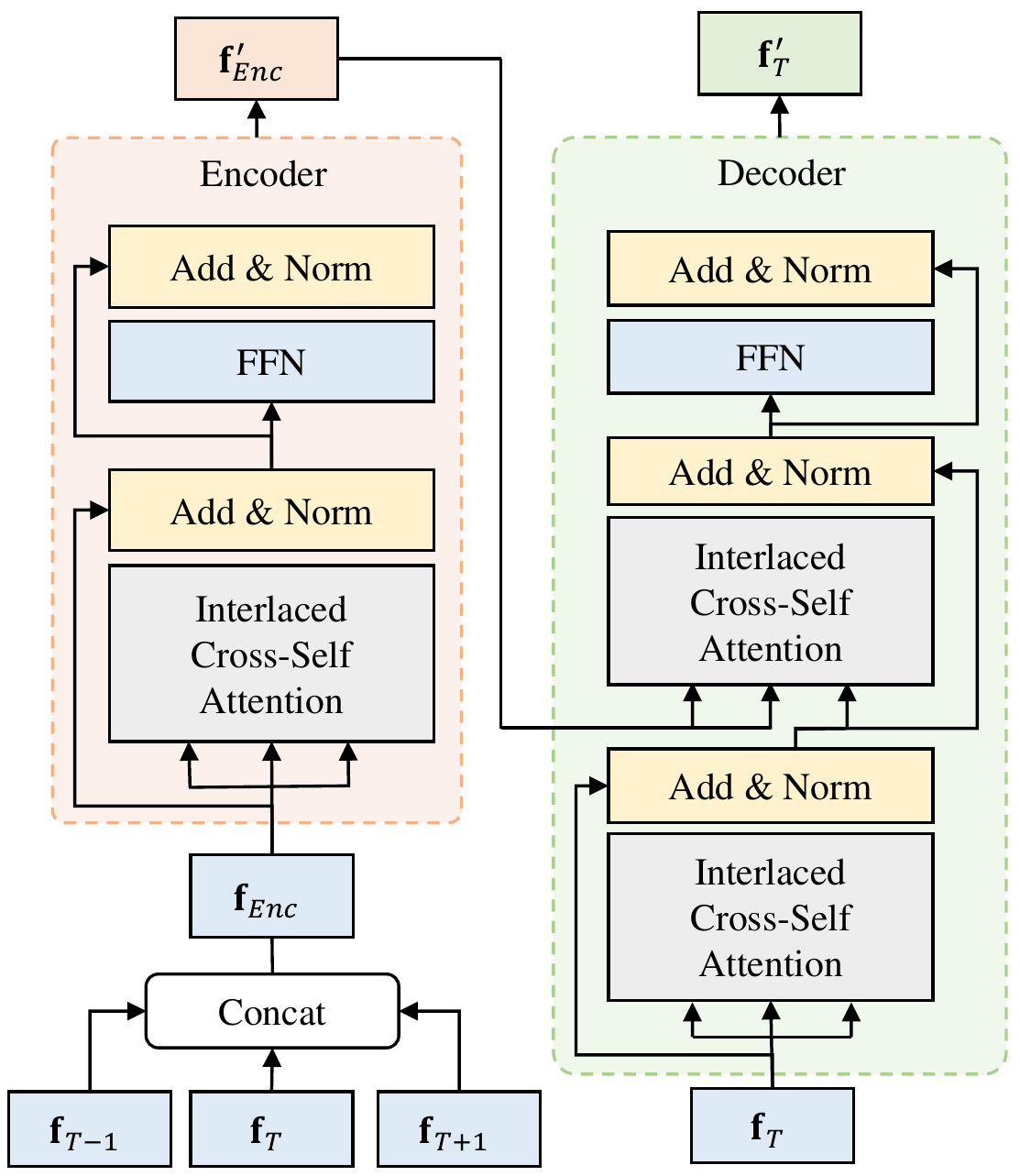}
	\end{center}
	\caption{\textbf{Illustration of transformer based spatial-temporal fusion module}. STF consists of an encoder for modeling spatial-temporal relationships and feature encoding, and a decoder for retrieving the feature of current frame from the encoded feature $\mathbf{f}_{Enc}^{'}$. Best viewed in color.}
	\label{transformer}
\end{figure}

In this work, we propose a spatial-temporal fusion module to effectively integrate the features of consecutive frames. Here it is expected that the spatial-temporal relationship among consecutive frames is well modeled and the optical flow estimation is avoided. In particular, we use the transformer~\cite{vaswani2017attention} to perform inter-frame fusion, which recently achieves the amazing performance in both NLP and CV areas. Thus our STF consists of an encoder and a decoder, as shown in Figure.~\ref{transformer}. 


\paragraph{Encoder} In STF, the encoder is used to capture the spatial-temporal relationships of pixel-level features. To this end, we concatenate the 2D features of multiple frames \{$\mathbf{f}_{T-1}$, $\mathbf{f}_{T}$, $\mathbf{f}_{T+1}$\} to obtain a 3D feature $\mathbf{f}_{Enc} \in \mathbb{R}^{d \times 3 \times H \times W}$, where $d$ is the dimension of pixel-level features, $H$ and $W$ represent the spatial size of frame features. That is, there are $3HW$ features in total for processing in the encoder. We first pass $\mathbf{f}_{Enc}$ into our proposed interlaced cross-self attention (ICSA) module to model dense spatial-temporal relationships, and the features are adjusted by weighting on all features. Then we feed the new features into feed-forward network (FFN) to perform feature transformation. Similar to~\cite{vaswani2017attention}, we employ the residual connections for the attention module and FFN followed by {layer normalization}. Finally, we obtain the encoded features $\mathbf{f}_{Enc}^{'}$.
Compared with the previous optical flow based  methods~\cite{gadde2017semantic,nilsson2018semantic}, our proposed STF uniformly aggregates all features at different spatial and temporal positions, and no explicit feature alignment is required. Essentially, a single feature in STF is implicitly aligned with multiple similar features by attention other than the temporally-aligned ones. This is reasonable since the purpose of feature fusion is to mutually enhance the features belonging to the same semantic class.

\paragraph{Decoder} In STF, the decoder is used to get the prediction features of the current frame. To this end, we use the original feature of current frame to retrieve from the encoded features $\mathbf{f}_{Enc}^{'}$. To be specific, we first feed the feature of current frame into an ICSA module to enhance the features similar in the encoder. 
Then we pass the enhanced features together with $\mathbf{f}_{Enc}^{'}$ into another ICSA module for cross attention and produce the features $\mathbf{f}_{T}^{'}$ with FFN. Different from the previous one, here the enhanced $\mathbf{f}_{T}$ serves as the query and $\mathbf{f}_{Enc}^{'}$ serves as the key and value. Intuitively, we retrieve the encoded features from $\mathbf{f}_{Enc}^{'}$ for each pixel-level feature in $\mathbf{f}_{T}$, and consequently the $\mathbf{f}_{T}^{'}$ would contain rich information from other spatial and temporal positions.

\begin{figure}[t]
	\begin{center}
		\includegraphics[width=0.8\linewidth]{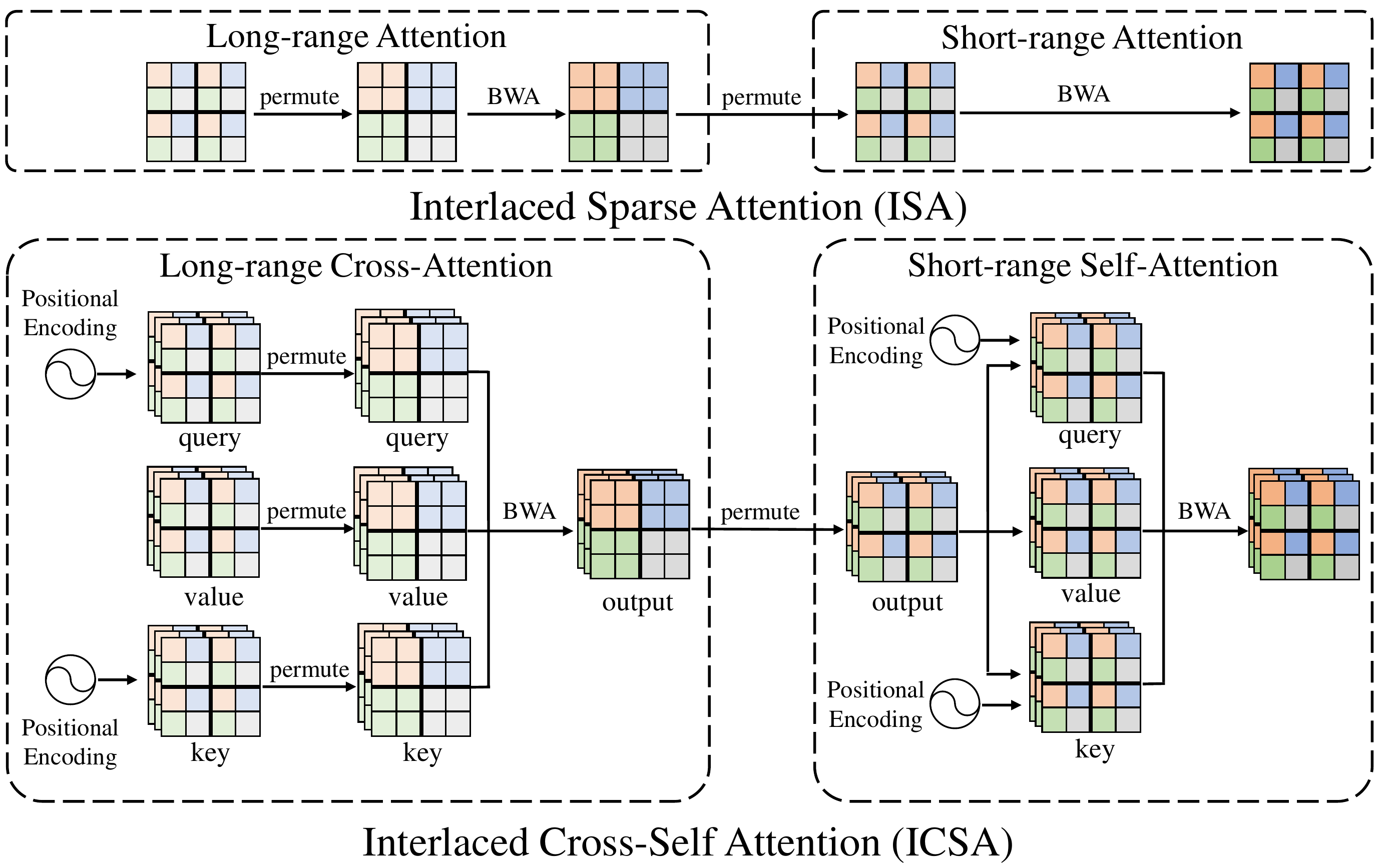}
	\end{center}
	\caption{\textbf{Illustration of differences between our interlaced cross-self attention (ICSA) and with interlaced sparse attention (ISA)~\cite{huang2019interlaced}}. ICSA takes query, key and value separately for long-rang cross-attention first and then conduct short-range self-attention on the previous enhanced feature, which can be seamlessly integrated in the transformer structure, especially for cross-attention module in the decoder. Besides, ICSA implements necessary positional encoding and can deal with features from multiple frames directly, which can uniformly model spatial-temporal relationships. Best viewed in color.}
	\label{attention_framework}
\end{figure}

\begin{figure}[t]
	\begin{center}
		\includegraphics[width=0.7\linewidth]{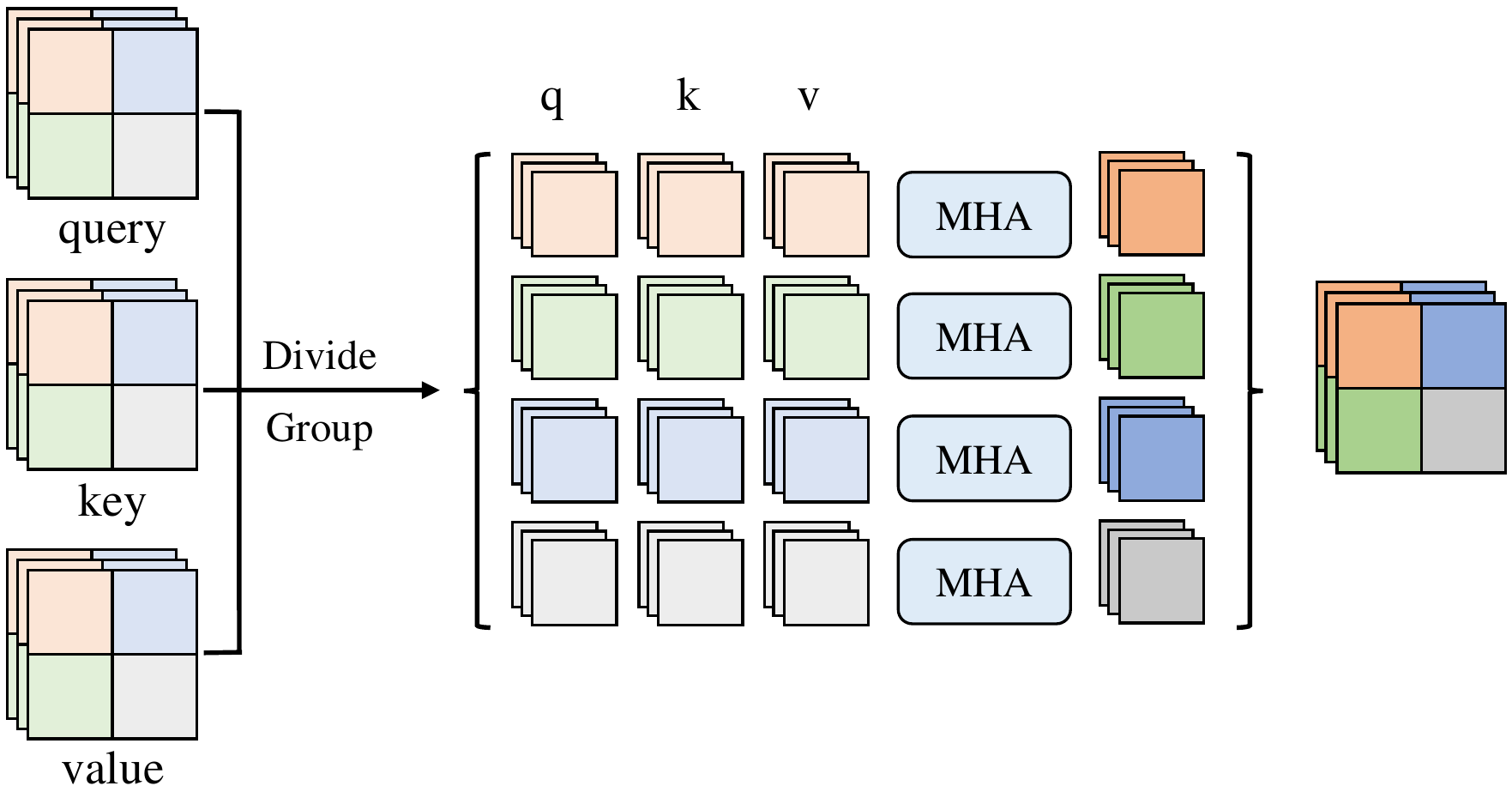}
	\end{center}
	\caption{\textbf{Illustration of block-wise attention (BWA)}. The input 3D features, \ie query, key and value, are spatially divided into patches with the same shape. Then we apply \textit{multi-head attention} (MHA)~\cite{vaswani2017attention} operation on corresponding query, key and value patches independently, and combine their results back to the entire one. Best viewed in color.}
	\label{block_wise}
\end{figure}

\paragraph{Interlaced Cross-Self Attention} In the original transformer, the attention operation would involve $\mathcal{O}(N^{2})$ complexity given an input of size $N$ (\eg, here $N=3HW$ in our case), which is impractical to the video semantic segmentation task since computation on pixel-level features would consume too much memory. 
To tackle this issue, a recent work ISA~\cite{huang2019interlaced} provides a successful solution. It decomposes the whole attention calculation as the combination of long-range and short-range sparse attention calculations, as shown in the upper subplot in Figure.~\ref{attention_framework}. In this way, it can retain the ability of modeling global relationship while effectively reduce the memory consumption. 
However, ISA is designed for self-attention mechanism like non-local~\cite{2017Non}, which is not well compatible with the transformer structure. Specifically, ISA takes a single feature as input and performs enhancement by modeling inner relationship. Thus it can not be directly integrated into cross attention in the transformer decoder. Besides, how to insert necessary positional encoding and deal with features of multiple frames are not considered by ISA.

In this work, we extend the original ISA into a more general form and propose interlaced cross-self attention (ICSA), which can be seamlessly integrated into transformer structure, as illustrated in the Figure.~\ref{attention_framework}. Generally, we reorganize ISA with long-range cross-attention and short-range self-attention operations. First, we take query, key and value separately as inputs for cross-attention.
Particularly, the query, key, and value are the same feature $\mathbf{f}_{Enc}$ for the STF module encoder, while the key and value are $\mathbf{f}_{Enc}^{'}$ and the query is the enhanced $\mathbf{f}_{T}$ for the STF module decoder. 
Here we directly takes 3D features as input to uniformly model spatial-temporal relationships.
For query and key, we supplement the features with positional encoding. Particularly, we choose the learnable positional encoding by following~\cite{vaswani2017attention}. In this work, we extend positional encoding to the 3D version and they have the same shape as the corresponding input.

Following ISA, we divide features into $k$ blocks with the same shape (\eg, $k=4$ in Figure.~\ref{attention_framework} and Figure.~\ref{block_wise}). To model long-range cross-attention, we harvest features with same spatial positions from different blocks in query, key and value via permutation operation, respectively. Then we conduct block-wise attention (BWA) operation for relationship modeling. As shown in Figure.~\ref{block_wise}, we first divide input query, key and value features into pre-defined blocks. Then, we apply multi-head attention (MHA)~\cite{vaswani2017attention} on corresponding query, key and value patches independently, and combine their results back to the entire one.
For short-range self-attention, we first permute the feature back to the original positions and then regard it as query, key and value for the next attention calculation. After adding positional encoding, we conduct BWA operation again and obtain the final enhanced feature.
With ICSA, STF can conveniently harvest global spatial-temporal information for feature enhancement while keeps an efficient attention computation.
Besides, if we take a single feature from one frame as query, key and value, and remove positional encoding, ICSA would degenerate into ISA. Evidently, ISA is a special case of ICSA.

\begin{figure}[t]
	\begin{center}
		\includegraphics[width=0.6\linewidth]{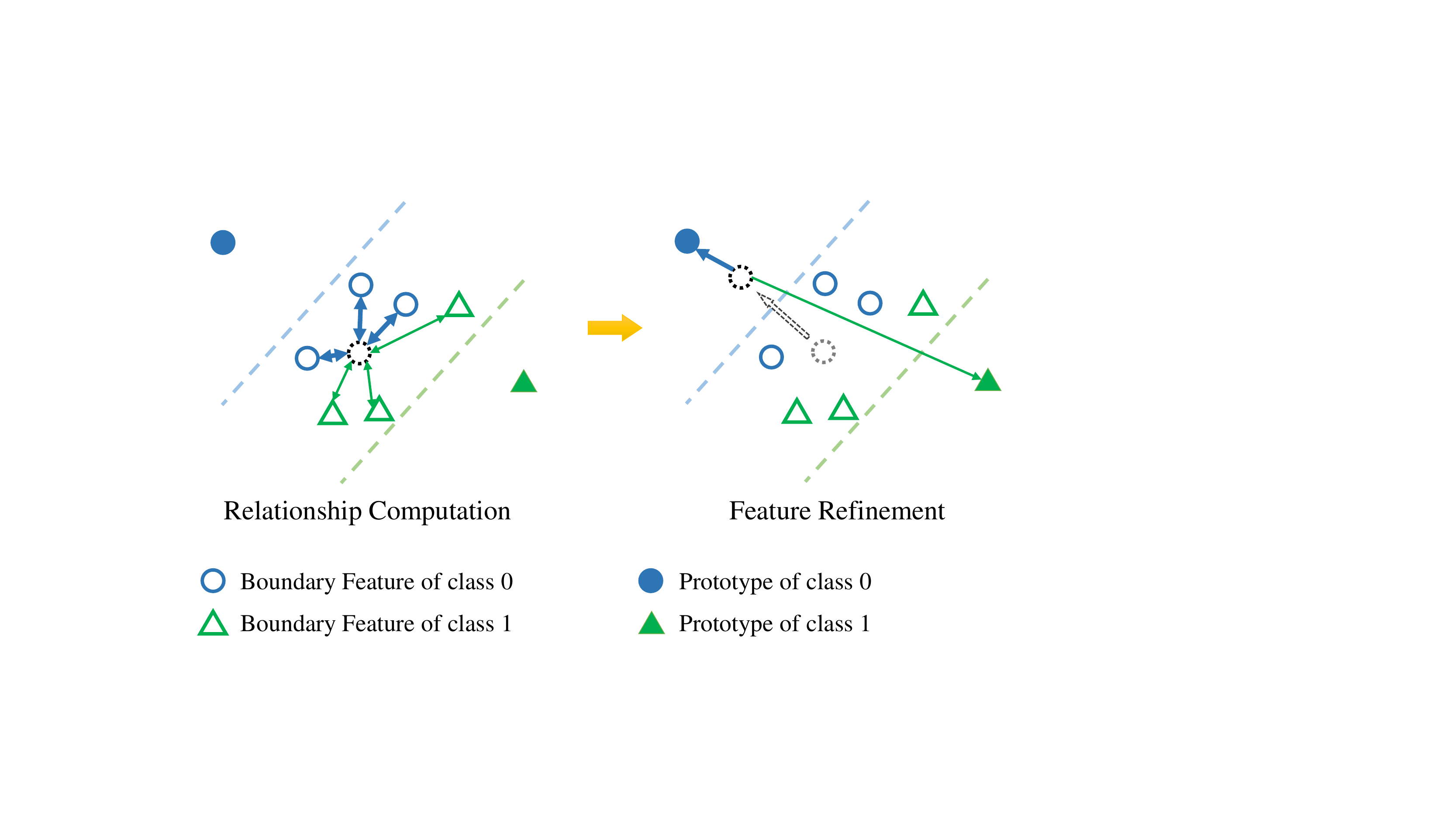}
	\end{center}
	\caption{\textbf{Illustration of feature refinement}. We first compute the relationships between the stored boundary features and the test feature to estimate the class likelihoods. Then we refine the test feature using the class prototypes, which essentially makes the feature move closer to the most likely class. Best viewed in color.}
	\label{MAR-idea}
\end{figure}

\subsection{Memory-Augmented Refinement}

In this work, we propose a novel memory-augmented refinement module to further refine the fused features. Different from previous works that explore the relationship among the inference features~\cite{krahenbuhl2011efficient,mazzini2018guided,yuan2020segfix}, we focus on refining the hard features (\eg, boundary features) using the memory from the training samples. The idea is illustrated in Figure.~\ref{MAR-idea}, and it is actually inspired by an intuitive mechanism of humans to process semantically ambiguous contents.
Specifically, given a test feature during inference, it usually lies in the boundary area of different classes in the feature space if it is hard to distinguish (\eg, with low confidence score). To enhance its discriminativeness, we first estimates its likelihoods to different classes by computing the similarities between the feature and stored boundary features of each class. Then we use the class prototypes to refine the feature according to the estimated likelihoods, where the class prototype refers to the mean feature representing a category. Through this way, the test feature would move closer to the most likely category.

\begin{figure}[t]
	\begin{center}
		\includegraphics[width=0.6\linewidth]{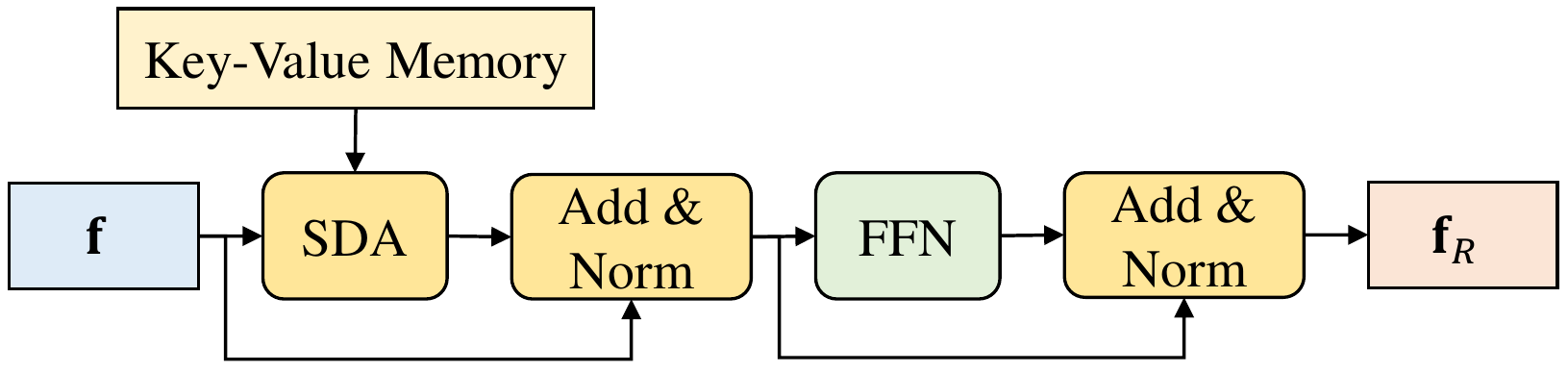}
	\end{center}
	\caption{\textbf{Illustration of memory-augmented refinement module}. The input feature $\mathbf{f}$ is refined into $\mathbf{f}_{R}$ using the key-value memory extracted from the training samples. Here $\mathbf{f}$ serves as the query, the key is the stored boundary features, and the value is the class prototypes. Here 'SDA' represents \textit{scaled dot-product attention} and 'FFN' represents \textit{feed forward network}. }
	\label{MARM}
\end{figure}

Our proposed MAR module is used to implement such an idea and is illustrated in Figure.~\ref{MARM}.
Specifically, we build the key-value memory for each class that stores two kinds of data from the training samples, namely, the boundary features and class prototypes. The boundary features serve as the keys $K \in \mathbb{R}^{d \times CK_{L}}$ and the class prototypes serve as the values $V \in \mathbb{R}^{d \times C}$, where $C$ denotes the number of classes and $K_{L}$ is a hyper-parameter to control the size of memory. 
In the MAR module, the input feature $F$ is refined into $F_{R}$ using the key-value memory.
Inspired by the transformer, we use the scaled dot-product attention (SDA) and FFN to construct the MAR block. To be specific, we take the test feature as query $Q \in \mathbb{R}^{d}$, and use the key-value in memory to refine it, resulting in $Q^{'}$. Formally, 
\begin{equation}
s_{i}=\theta(Q)^{T}\phi(K_{i}),
\end{equation}
\begin{equation}
s_{i} = \frac{e^{s_{i}}}{\sum_{i=1}^{CK_{L}}e^{s_{i}}},
\end{equation}
\begin{equation}
Q^{'} = \sum_{i=1}^{CK_{L}}s_{i}V_{j},
\label{Eq.3}
\end{equation}
where $i \in [1,CK_{L}]$ denotes the sample in memory and $j$ is the class index corresponding to the ${i}$-th sample. Here $\theta(Q)=W_{\theta}Q$ and $\phi(K_{i})=W_{\phi}K_{i}$, and $W_{\theta}$ and $W_{\phi}$ are two learnable matrices. Notably, in Eq.~(\ref{Eq.3}), we index $V$ by $j$ rather than $i$, which is different from the original self-attention calculation. We employ the residual connections for SDA and FFN followed by layer normalization, like in the original transformer~\cite{vaswani2017attention}.

\begin{figure}[t]
	\begin{center}
		\includegraphics[width=0.8\linewidth]{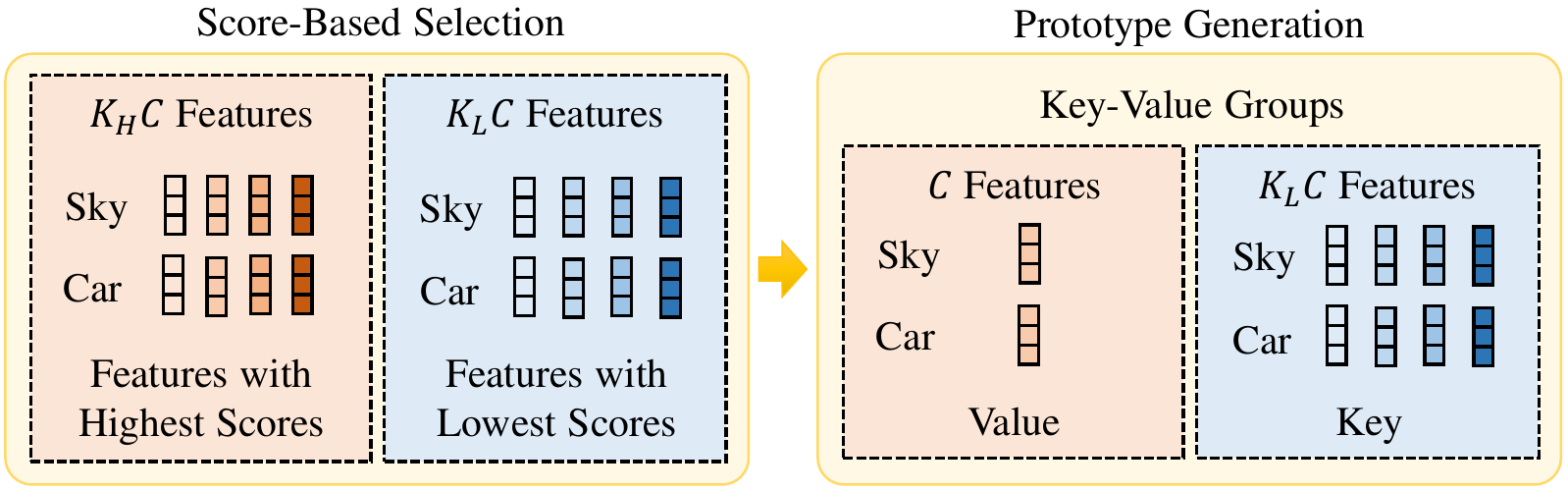}
	\end{center}
	\caption{\textbf{Illustration of the key-value memory}. From the extracted features on the training set, we select $K_H$ "good" features with the highest scores and $K_{L}$ "hard" features with the lowest scores per class. Then we generate the class prototypes by averaging the "good" features, and organize them with "hard" features to form the key-value memory. Here $K_{H}=4$, $K_{L}=4$, and $C=2$ are used for visualization. }
	\label{memory}
\end{figure}

Next we explain how to generate the key-value memory from the training samples, which is shown in Figure.~\ref{memory}. 
We first train the segmentation network without the MAR module. Using this model, we extract the features for all training samples. Note that a feature would be discarded if it is misclassified by the classifier. According to the ground truth, for each class, we select $K_L$ "hard" features with the lowest confidence scores and $K_H$ "good" features with the highest confidence scores. The former are considered to suffer from semantically ambiguity while the latter are to accurately represent the semantic category. After that, we compute the mean feature of the "good" features for each class, resulting in the class prototype. Finally, we store the "hard" features as keys and the corresponding class prototype as values in the memory, which essentially represent the task-specific experience.

\subsection{Training Strategy}

Our proposed network consists of four main components, \ie, backbone, classifier, STF, and MAR. Here, we adopt a multi-stage training schedule, which is a common strategy in advanced works, \eg, Faster RCNN and knowledge distillation. First, the backbone and classifier together are pretrained on ImageNet and finetuned on a particular segmentation dataset (\eg, Cityscapes and CamVid). The backbone would keep fixed and the classifier would be re-initialized in the following training procedures. Then, we train STF together with the backbone and classifier, and use this model to generate the key-value memory. Finally, we train the MAR and classifier by fixing STF. In the test phase, we perform the end-to-end inference with STF and MAR.

\section{Experiment}
\label{sec:exp}

In this section, we experimentally evaluate our proposed method on two challenging datasets by following the previous works, namely, Cityscapes~\cite{cordts2016cityscapes} and CamVid~\cite{brostow2009semantic}. Here some state-of-the-art methods are adopted for comparison and the results from the literatures are listed. We follow the standard protocols of video semantic segmentation and report the mean Intersection over Union (mIoU) as the performance metric.

\subsection{Dataset}
\textbf{Cityscapes}~\cite{cordts2016cityscapes} is a popular dataset in semantic segmentation and autonomous driving domain. It focuses on semantic understanding of urban street scenes. The training and validation subsets contain $2,975$ and $500$ video clips, respectively, and each video clip contains $30$ frames. The $20${th} frame in each clip is annotated by the pixel-level semantic labels with $19$ categories.

\textbf{CamVid}~\cite{brostow2009semantic} also focuses on the semantic understanding of urban street scenes, but it contains less data than Cityscapes. It only has $701$ color images with annotations of $11$ semantic classes. CamVid is divided into the trainval set with $468$ samples and test set with $233$ samples.  All samples are extracted from the driving videos captured at daytime and dusk, and have pixel-level semantic annotations. Each CamVid video contains $3,600$ to $11,000$ frames at a resolution of $720\times960$.

\subsection{Performance Comparison}

\begin{table}[t]
	\footnotesize
	\begin{center}
		\begin{tabular}{l|c|c}
			\toprule
			\multicolumn{1}{c|}{Method}		& Backbone	& mIoU (\%)	   				\\
			\hline\hline
			PSPNet~\cite{zhao2017pyramid} 		& ResNet18	& 69.79						\\
			+ Liu~\etal~\cite{liu2020efficient}	& ResNet18	& 73.06	(+3.27)				\\
			+ \textbf{Ours} 					& ResNet18	& \textbf{74.58 (+4.79)}	\\
			\hline\hline
			PSPNet~\cite{zhao2017pyramid} 		& ResNet50	& 76.24						\\
			+ Accel~\cite{jain2019accel} 		& ResNet50	& 70.20	(-6.04)				\\
			+ TDNet~\cite{hu2020temporally} 	& ResNet50	& 76.40	(+0.27)				\\
			+ EFC~\cite{ding2020every} 			& ResNet50	& 78.44	(+2.31)				\\
			+ \textbf{Ours} 					& ResNet50	& \textbf{79.22 (+2.98)}	\\
			\hline \hline
			PSPNet~\cite{zhao2017pyramid}		& ResNet101	& 79.70						\\
			+ TDNet~\cite{hu2020temporally} 	& ResNet101	& 79.90	(+0.20)				\\
			+ NetWarp~\cite{gadde2017semantic} 	& ResNet101	& 80.60 (+0.90)				\\
			+ GRFP~\cite{nilsson2018semantic} 	& ResNet101	& 80.20 (+0.50)				\\			
			+ \textbf{Ours} 					& ResNet101	& \textbf{80.96 (+1.26)}	\\
			\hline \hline
			Swin-B~\cite{liu2021swin}			& Swin Transformer	& 81.34						\\
			+ \textbf{Ours} 					& Swin Transformer	& \textbf{81.67 (+0.33)}	\\
			\bottomrule
		\end{tabular}
	\end{center}
	\caption{Performance comparison on Cityscapes val subset. PSPNet and Swin Transformer are chosen as the image segmentation models.}
	\label{cityscapes}
\end{table}

\begin{table}[t]
	\footnotesize
	\begin{center}
		\begin{tabular}{l|c|c}
			\toprule
			\multicolumn{1}{c|}{Method}				& Backbone	& mIoU (\%)	   				\\
			\hline\hline
	        Dilation8~\cite{yu2015multi}  			& VGG16		& 65.3						\\
			+ STFCN~\cite{fayyaz2016stfcn} 			& VGG16		& 65.9 (+0.4)				\\
			+ GRFP~\cite{nilsson2018semantic} 		& VGG16		& 66.1 (+0.8)				\\
			+ FSO~\cite{kundu2016feature} 			& VGG16		& 66.1 (+0.8)				\\
			+ VPN~\cite{jampani2017video} 			& VGG16		& 66.7 (+1.4)				\\
			+ NetWarp~\cite{gadde2017semantic} 		& VGG16		& 67.1 (+1.8)				\\
			+ EFC~\cite{ding2020every} 				& VGG16		& 67.4 (+2.1)				\\
			+ \textbf{Ours} 						& VGG16		& \textbf{67.9 (+2.6)}		\\
			\hline\hline
			PSPNet~\cite{zhao2017pyramid}			& ResNet101	& 76.2						\\
			+ Accel~\cite{jain2019accel}			& ResNet101	& 71.5 (-4.7)				\\
			+ TDNet~\cite{hu2020temporally}			& ResNet101	& 76.0 (-0.2)				\\
			+ \textbf{Ours}							& ResNet101	& \textbf{76.6 (+0.4)}		\\
			\hline\hline
			Swin-B~\cite{liu2021swin}				& Swin Transformer	& 77.6						\\
			+ \textbf{Ours}							& Swin Transformer	& \textbf{77.9 (+0.3)}		\\
			\bottomrule
		\end{tabular}
	\end{center}
	\caption{Performance comparison on CamVid test subset. Dilation8, PSPNet and Swin Transformer are chosen as the image segmentation model.}
	\label{camvid}
\end{table}

Here we compare our proposed method with the state-of-the-art methods on Cityscapes and CamVid. In particular, the image segmentation model is used as the baseline. The PSPNet~\cite{zhao2017pyramid} with the backbone ResNet18/50/101 has been widely used on Cityscapes, and Dilation8~\cite{yu2015multi} is mainly adopted on CamVid.
Table~\ref{cityscapes} and Table~\ref{camvid} show the results, and we have the following observations. 
First, our proposed method achieves the state-of-the-art performance on both datasets and various baseline model, which demonstrate the effectiveness and generalization of our method.
Second, our proposed method can get more gains on light-weight baseline model. This is reasonable since improving more complicated model is generally more difficult.
Third, TDNet~\cite{hu2020temporally}, Accel~\cite{jain2019accel} and Liu~\etal~\cite{liu2020efficient} have nearly no improvement and even degradation on accuracy comparing to the baseline, since they mainly focus on improving inference speed.
Fourth, even on the strong baseline, \eg, Swin Transformer~\cite{liu2021swin}, our method can also bring improvement.

\subsection{Ablation Study}

\paragraph{Effectiveness of our method}

\begin{table}[t]
	\footnotesize
	\begin{center}
		\begin{tabular}{l|c|c|c}
			\toprule
			\multicolumn{1}{c|}{Method}			& Dataset		& Backbone		& mIoU (\%)	   				\\
			\hline\hline
			PSPNet~\cite{zhao2017pyramid} 		& Cityscapes	& ResNet50		& 76.24						\\
			+ STF 								& Cityscapes	& ResNet50		& 78.75 (+2.62)				\\
			+ MAR 								& Cityscapes	& ResNet50		& 78.37 (+2.24)				\\
			+ \textbf{Both} 					& Cityscapes	& ResNet50		& \textbf{79.22 (+2.98)}	\\
			\hline\hline
			Dilation8~\cite{yu2015multi} 		& CamVid		& VGG16			& 65.3						\\
			+ STF 								& CamVid		& VGG16			& 67.5 (+2.2)				\\
			+ MAR 								& CamVid		& VGG16			& 67.3 (+2.0)				\\
			+ \textbf{Both} 					& CamVid		& VGG16			& \textbf{67.9 (+2.6)}		\\
			\bottomrule
		\end{tabular}
	\end{center}
	\caption{Ablation study on key modules of our proposed method. Performance comparison on Cityscapes val subset and CamVid test subset. PSPNet and Dilation8 are chosen as the image segmentation model, respectively.}
	\label{effectiveness}
\end{table}

In this work, we propose two key modules, namely STF and MAR. To investigate their effects, we also give the results of applying one of them, as shown in Table~\ref{effectiveness}. We can have the following observations. First, our proposed STF and MAR can bring significant performance improvement separately compared with the baseline, and the version equipped with both of them performs best. 
Second, STF can brings more gains than MAR, since STF integrates the multi-frame information while MAR only optimizes the features within the current frame.
Third, our proposed STF outperforms other multi-frame fusion methods (\eg, TDNet~\cite{hu2020temporally} in Table~\ref{cityscapes} and GRFP~\cite{nilsson2018semantic} and NetWarp~\cite{gadde2017semantic} in Table~\ref{camvid}), which indicates the effectiveness of modeling spatial-temporal relationship.

\begin{table}[t]
	\footnotesize
	\begin{center}
		\begin{tabular}{l|c}
			\toprule
			\multicolumn{1}{c|}{Method}					& mIoU (\%)					\\			
			\hline\hline
			DeepLabv3~\cite{chen2017rethinking}    		& 79.5						\\
			+ DenseCRF~\cite{krahenbuhl2011efficient} 	& 79.7 (+0.2)				\\
			+ GUM~\cite{mazzini2018guided} 				& 79.8 (+0.3)				\\
			+ SegFix~\cite{yuan2020segfix}				& 80.5 (+1.0)				\\
			+ \textbf{Our MAR} 							& \textbf{81.0} (+1.5)		\\
			\bottomrule
		\end{tabular}
	\end{center}
	\caption{Comparison of different feature refinement methods on Cityscapes val subset. DeepLabv3 is chosen as the baseline model.}
	\label{MAR}
	\vspace{-2mm}
\end{table}

\begin{figure*}[t]
	\begin{center}
		\includegraphics[width=1.0\linewidth]{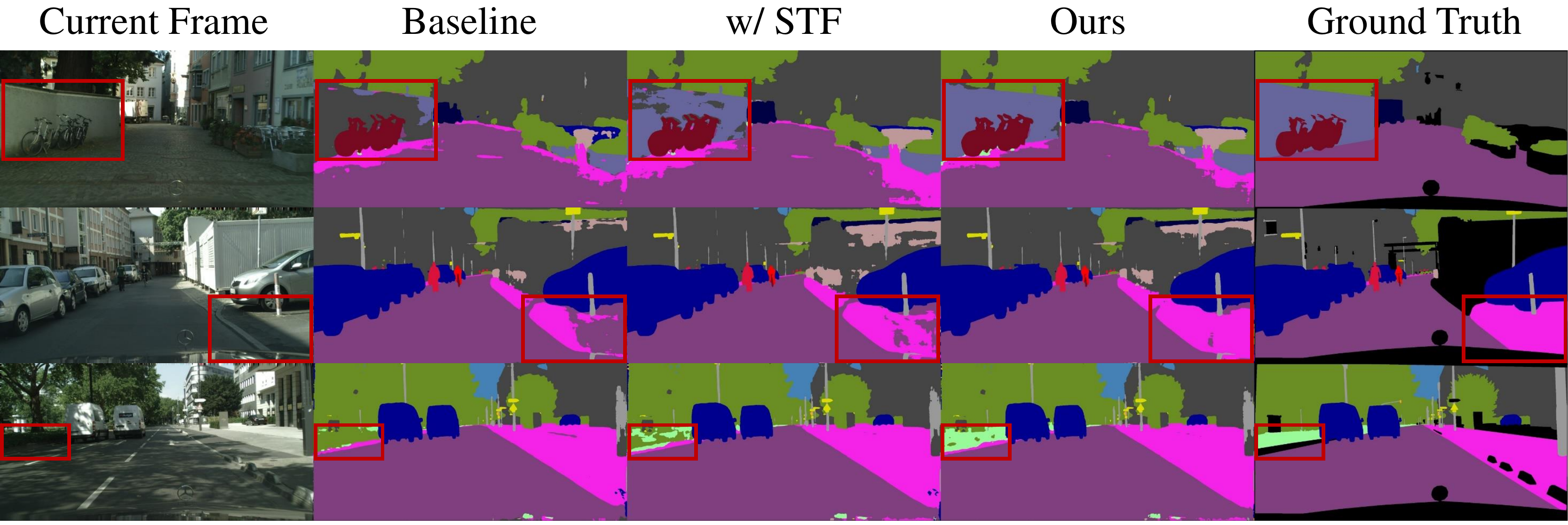}
	\end{center}
	\caption{\textbf{Visualization of some sample segmentation results from Cityscapes}. It can be seen that STF can significantly improve the baseline results, and MAR can further bring gains. Here the red rectangles highlight the important regions. Best viewed in color.}
	\label{vis}
\end{figure*}

\paragraph{Analysis of MAR}

In this paper, we propose MAR to refine the video features. Actually, this technique can also be used in other tasks. Here we particularly investigate its effect on image segmentation, and show its superiority by comparing with other representative refinement methods, including DenseCRF~\cite{krahenbuhl2011efficient}, GUM~\cite{mazzini2018guided}, and SegFix~\cite{yuan2020segfix}.  DenseCRF~\cite{krahenbuhl2011efficient} establishes pairwise potentials on all pairs of pixels and poses segmentation refinement problem as maximum a posteriori (MAP) inference, which is a classic post-processing method.
GUM~\cite{mazzini2018guided} proposes to enrich bilinear upsampling operators by introducing a learnable transformation for semantic maps, which can steer the sampling towards the correct semantic class.
SegFix~\cite{yuan2020segfix} proposes to replace the unreliable predictions of boundary pixels with the predictions of interior pixels, which currently achieves the state-of-the-art performance.
Table~\ref{MAR} provides the comparison results, where the DeepLabv3 is adopted as the baseline by following previous works. We can see that our MAR outperforms the previous methods, which indicates the effectiveness of refining feature by memory.

\begin{figure}[t]
	\begin{center}
		\includegraphics[width=0.8\linewidth]{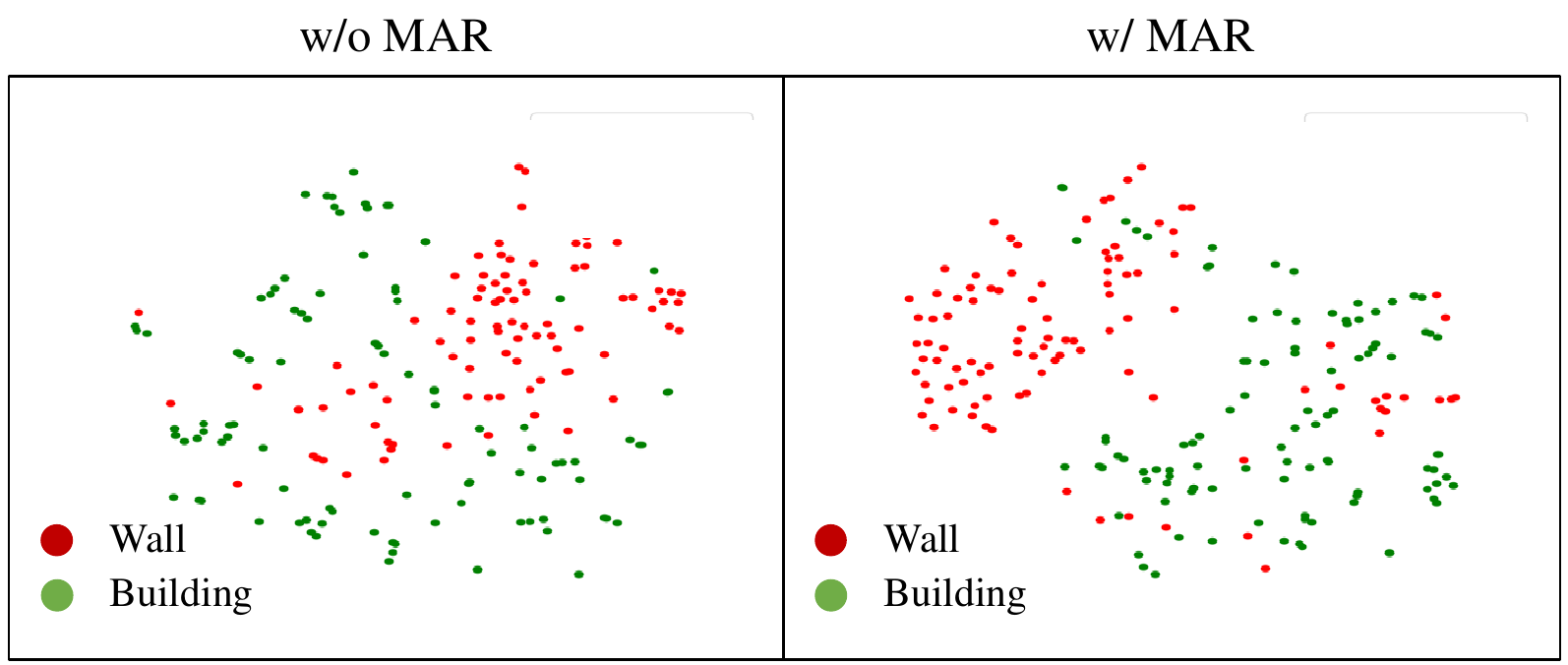}
	\end{center}
	\caption{\textbf{Visualization on the change of feature distribution}. It can be seen that features become more separable after using MAR module. Best viewed in color.}
	\label{vis_MAR}
\end{figure}

To intuitively show the effect of MAR on feature refinement, we particularly choose two easily ambiguous categories, \ie, wall and building, to visualize the features before and after applying MAR. To be specific,  we randomly sample $100$ hard features (with confidence scores lower than 0.8) per category, and then visualize their distribution using t-SNE. The results are shown in Figure.~\ref{vis_MAR}. Before using MAR, two kinds of features are confused together. MAR can move features closer to their corresponding class prototypes and make them easier to be separated.

\paragraph{Hyper-parameter $K_{L}$ and $K_{H}$}

In our MAR, $K_{L}$ and $K_{H}$ are used to control the number of boundary features for memory and good features for prototype per class. Here we explore their influence on the segmentation accuracy. Considering the memory size, we particularly evaluate the $K_{L}$ from $\{10, 50, 100, 300\}$ and the $K_{H}$ from $\{1, 5, 10, 50\}$ on Cityscapes val subset with PSPNet-ResNet50 as the base model. We found that there is almost no performance fluctuation for different settings. Finally, $K_{L}=10$ and $K_{H}=10$ are adopted throughout the experiments. 

\paragraph{Analysis of segmentation results}

\begin{table*}[t]
	\footnotesize
	\begin{center}
		\setlength{\tabcolsep}{1.5mm}
		\begin{tabular}{l|cccccccccc}
			\hline
			\multicolumn{1}{c|}{Method}      & road	& side.	& build.	& wall	& fence	& pole	& light	& sign	& vege.	& terr.	\\
			\hline\hline
			PSPNet-50		& 97.82	& 83.23 & 91.70 & \textbf{35.86} & 58.07 & 63.87 & 70.76 & 78.93 & 91.95 & \textbf{62.85}			\\
			+ MAR		& 98.13	& 85.02 & 92.39 & \textbf{51.34} & 60.81 & 63.46 & 71.62 & 80.49 & 92.55 & \textbf{65.47}			\\
			\hline
			\multicolumn{1}{c|}{Method}		& sky	& pers.	& rider	& car	& truck	& bus	& train	& motor	& bike	& mean	   							\\
			\hline\hline
			PSPNet-50		& 94.16 & 81.86 & 60.96 & 94.82 & \textbf{76.34} & \textbf{85.83} & 77.67 & \textbf{64.43} & 77.61 & 76.24	\\
			+ MAR		& 94.52 & 82.42 & 60.88 & 95.34 & \textbf{80.72} & \textbf{88.72} & 78.73 & \textbf{68.28} & 78.21 & 78.37	\\
			\hline
		\end{tabular}
	\end{center}
	\caption{Category-wise performance on Cityscapes val subset. PSPNet-50 is chosen as the baseline model.}
	\label{compare}
\end{table*}

Our proposed MAR mainly handles the ambiguous cases, especially for the hard classes. Table~\ref{compare} lists the segmentation accuracy of different semantic classes on Cityscapes, where PSPNet with the  ResNet50 backbone is particularly adopted as the baseline. We can see that our method can consistently boost the accuracy over all classes and the gain is especially significant for the hard classes, \eg, wall, terrain, and truck.
In addition, to intuitively show the effectiveness of our proposed STF and MAR, we visualize three sample segmentation results from Cityscapes in Figure.~\ref{vis}. It can be seen that the original segmentation results can be progressively improved by STF and MAR.

\paragraph{Cost of STF and MAR}

\begin{table}[t]
	\footnotesize
	\begin{center}
		\begin{tabular}{c|c}
			\toprule
			Module          		& MACs (G)	\\
			\hline
			PSPNet-50 (3 Frames) 	& 4285.9	\\
			PSPNet-101 (3 Frames) 	& 6152.9	\\
			STF 					& 563.5		\\
			MAR 					& 31.5		\\
			Classifier 				& 0.5		\\
			\bottomrule
		\end{tabular}
	\end{center}
	\caption{Computational cost of different modules (GMACs). The resolution of input image is $1024 \times 2048$.}
	\label{computation}
\end{table}

Here we analyze the computational cost of different components in our proposed method, and the statistics are provided in Table~\ref{computation}. It can be seen that our STF and MAR involve little computational cost compared with the base model. In particular, our proposed MAR is more efficient to achieve good segmentation performance than devising more complicated network structures.

\section{Conclusion} \label{sec:conclusion}
In this paper, we design a novel video semantic segmentation framework with inter-frame feature fusion and inner-frame feature refinement, and propose two novel techniques to boost the accuracy.
Specifically, we first propose a spatial-temporal fusion module based on the transformer, which can effectively aggregate multi-frame features at different spatial and temporal positions, and meanwhile avoid error-prone optical flow estimation. Then we propose a memory-augmented refinement module that exploits the stored features from the training samples to augment the hard features during inference. Our experimental results on Cityscapes and CamVid show that the proposed method outperforms the state-of-the-art methods.

\section*{Acknowledgements}
This work is supported by the National Natural Science Foundation of China under Grant No.62176246 and No.61836008.
We acknowledge the support of GPU cluster built by MCC Lab of Information Science and Technology Institution, USTC.









\bibliographystyle{elsarticle-num} 
\bibliography{egbib}

\end{document}